\pdfoutput=1

\documentclass[11pt]{article}

\usepackage{emnlp2021}

\usepackage{times}
\usepackage{latexsym}
\usepackage{inconsolata}

\usepackage[T1]{fontenc}
\usepackage[utf8]{inputenc}

\usepackage{microtype}

\usepackage{amsmath}
\usepackage{amsfonts}
\usepackage{amsthm}
\usepackage{amssymb}
\usepackage{bbm}
\usepackage{booktabs}
\usepackage{longtable}
\usepackage{float}
\usepackage{multirow}
\usepackage{xspace}
\usepackage{xfrac}
\usepackage{enumitem}

\usepackage{graphicx} 
\graphicspath{{./figs/}} 

\usepackage{caption}
\usepackage{subcaption}

\usepackage{pifont}
\newcommand{\cmark}{\ding{51}}
\newcommand{\xmark}{\ding{55}}

\usepackage{xcolor}

\usepackage[disable]{todonotes}
\makeatletter
\newcommand*\iftodonotes{\if@todonotes@disabled\expandafter\@secondoftwo\else\expandafter\@firstoftwo\fi}  %
\makeatother

\newcommand{\note}[4][]{\todo[author=#2,color=#3,size=\scriptsize,fancyline,caption={},#1]{#4}} %

\newcommand{\tiago}[2][]{\note[#1]{tiago}{cyan!40}{#2}}

\newcommand{\clara}[2][]{\note[#1]{clara}{orange!40}{#2}}

\newcommand{\defn}[1]{\textbf{#1}}

\newcommand{\normal}{\mathcal{N}}
\newcommand{\R}{\mathbbm{R}}
\newcommand{\softmax}{\mathrm{softmax}}
\newcommand{\lstm}{\mathrm{LSTM}}

\newcommand{\ent}{\mathrm{H}}

\newcommand{\btheta}{{\boldsymbol \theta}}
\newcommand{\ptheta}{p_{\btheta}}
\newcommand{\enttheta}{\ent_{\btheta}}

\newcommand{\bs}{\boldsymbol{s}}
\newcommand{\bS}{\boldsymbol{S}}
\newcommand{\be}{\mathbf{e}}
\newcommand{\bh}{\mathbf{h}}
\newcommand{\bb}{\mathbf{b}}
\newcommand{\bx}{\mathbf{x}}
\newcommand{\bW}{\mathbf{W}}
\newcommand{\calS}{\mathcal{S}}
\newcommand{\lang}{\ell_i}

\newcommand{\bphi}{\boldsymbol{\phi}}
\newcommand{\bbeta}{\boldsymbol{\beta}}

\renewcommand{\hat}{\widehat}

\usepackage{cleveref}
\crefname{section}{\S}{\S\S}
\Crefname{section}{\S}{\S\S}
\crefname{table}{Tab.}{}
\crefname{figure}{Fig.}{Figs.}
\crefname{algorithm}{Algorithm}{}
\crefname{algorithm}{Algorithm}{}
\crefname{line}{Line}{}
\crefname{appendix}{App.}{}
\crefformat{section}{\S#2#1#3}
\crefname{thm}{Theorem}{}
\crefname{cor}{Corollary}{}
\crefname{prop}{Proposition}{}
\crefname{def}{Definition}{}

\title{A surprisal--duration trade-off across and within the world's languages}

\usepackage{tipa}

\newcommand{\ucambridge}{\normalfont \text{\textipa{D}}}
\newcommand{\jhu}{\normalfont \text{\textipa{4}}}
\newcommand{\ethz}{\text{\normalfont \textipa{Q}}}
\newcommand{\harvard}{\normalfont \text{\textipa{@}}}
\newcommand{\mpi}{\normalfont \text{\textipa{T}}}
\newcommand{\hse}{\normalfont \text{\textipa{R}}}

\author{Tiago Pimentel$^{\ucambridge}$~\;~ Clara Meister$^{\ethz}$~\;~ Elizabeth Salesky$^{\jhu}$~\;~Simone Teufel$^{\ucambridge}$\\
\textbf{Dami\'an Blasi$^{\harvard,\mpi,\hse}$~\;~ Ryan Cotterell$^{\ucambridge,\ethz}$} \\
  $^{\ucambridge}$University of Cambridge~\;~%
  $^{\ethz}$ETH Z\"{u}rich~\;~%
  $^{\jhu}$Johns Hopkins University~\;~%
  $^{\harvard}$Harvard University \\
  $^{\mpi}$Max Planck Institute for the Science of Human History~\;~%
  $^{\hse}$Higher School of Economics \\
  \href{mailto:tp472@cam.ac.uk}{\texttt{tp472@cam.ac.uk}}~\;~%
  \href{mailto:clara.meister@inf.ethz.ch}{\texttt{clara.meister@inf.ethz.ch}}~\;~%
  \href{mailto:esalesky@jhu.edu}{\texttt{esalesky@jhu.edu}} \\
 \href{mailto:sht25@cl.cam.ac.uk}{\texttt{sht25@cl.cam.ac.uk}}~\;~\href{mailto:dblasi@fas.harvard.edu}{\texttt{dblasi@fas.harvard.edu}}~\;~ \href{mailto:ryan.cotterell@inf.ethz.ch}{\texttt{ryan.cotterell@inf.ethz.ch}}
}

\date{}

\begin{document}
\maketitle
\begin{abstract}
While there exist scores of natural languages, each with its unique features and idiosyncrasies, they all share a unifying theme: enabling human communication.
We may thus reasonably predict that human cognition shapes how these languages evolve and are used.
Assuming that the capacity to process information is roughly constant across human populations, we expect a surprisal--duration trade-off to arise both across and within languages.
We analyse this trade-off using a corpus of 600 languages and, after controlling for several potential confounds, we find strong supporting evidence in both settings.
Specifically, we find that, on average, phones are produced faster in languages where they are less surprising, and vice versa.
Further, we confirm that more surprising phones are longer, on average, in 319 languages out of the 600.
We thus conclude that there is strong evidence of a surprisal--duration trade-off in operation, both across and within the world's languages.
\end{abstract}

\section{Introduction}

During the course of human evolution, countless languages have evolved, each with unique features.
Despite their stark differences, however, it is plausible that shared attributes in human cognition may have placed constraints on how each language is implemented.
These constraints, in turn, may lead to compensations and trade-offs in the world's languages.
For instance, if we assume a channel capacity \citep{shannon1948mathematical} in human's ability to process language \citep[as posited by][]{frank2008speaking}, we may make predictions about these trade-offs.
Additionally, if we assume this capacity to be uniform across human populations, these trade-offs will extend cross-linguistically.\looseness=-1

\begin{figure}
    \centering
    \includegraphics[width=\columnwidth]{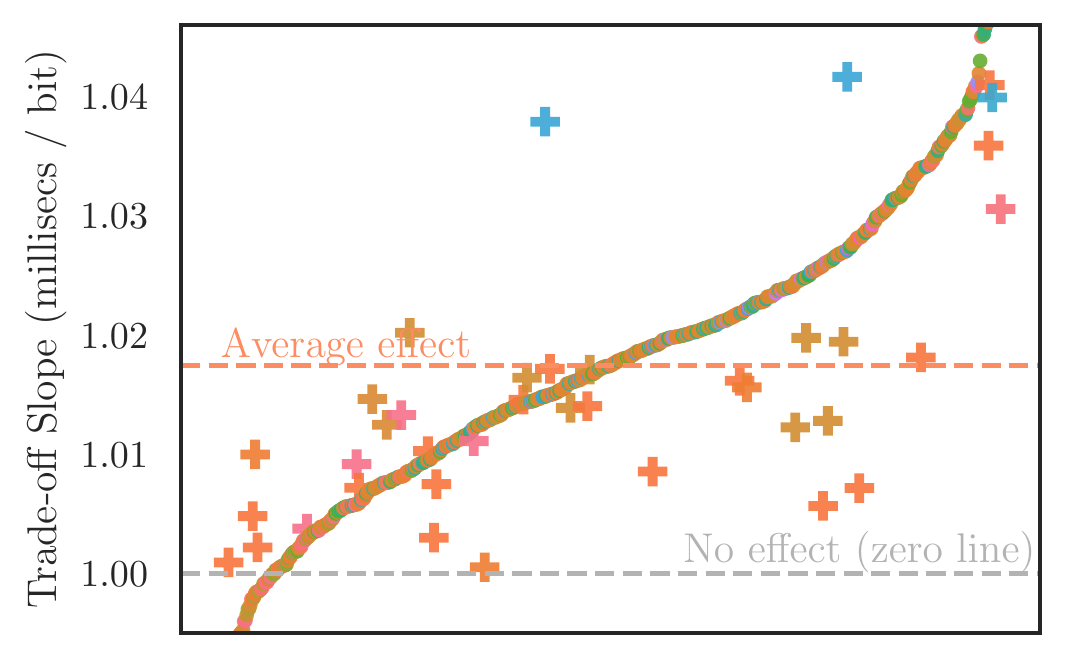}
    \caption{
    Surprisal--duration trade-off slopes. The $y$-axis presents a multiplicative effect, duration is multiplied by $y$ per bit of information.
    Sorted dots represent languages in Unitran; `+' are languages in Epitran.
    }
    \label{fig:monoling_effect_full}
\end{figure}

Within languages, there is a direct connection between this channel capacity assumption and the uniform information density hypothesis \citep[UID;][]{fenk1980konstanz,aylett2004smooth,levy2007speakers}, which predicts that speakers smooth the information rate in a linguistic signal so as to keep it roughly constant; by smoothing their information rate, natural languages can stay close to a (hypothetical) channel capacity.
Across languages, a unified channel capacity allows us to derive a specific instantiation of the compensation hypothesis \citep{hockett1958course}, with information density (measured in, e.g., bits per phone) being compensated by utterance speed (in, e.g., milliseconds per phone).
We may thus predict a trade-off between surprisal\footnote{Surprisal is defined as the negative log-probability of an event, e.g. observing a phone given its prior context.} and duration both within and across the world's languages.\looseness=-1

This trade-off has been studied amply within high resource languages \citep[][\emph{inter alia}]{genzel-charniak-2002-entropy,bell2003effects,mahowald2018word}. 
Cross-linguistically, however, this trade-off has received comparatively little attention, with a few notable exceptions such as \citet{pellegrino2011cross} and \citet{coupe2019different}.
Several factors have inhibited  cross-linguistic studies of this kind.
Arguably, the most prominent 
is the sheer lack of data necessary to investigate the phenomenon.
While massively cross-linguistic data abounds in the form of wordlists \citep{wichmann2020asjp,northeuralex}, surprisal is a context-dependent measure and, therefore, isolated word types are not enough for this analysis.
Further, as we have a specific hypothesis for why this trade-off should arise (humans' information processing capacity), we are not interested in simply finding \emph{any} correlation between surprisal and duration.
Several confounds could drive such a correlation, but most of these are either trivially true or uninteresting from our perspective.
Therefore, a thorough analysis of this trade-off needs to control for these potential confounds.\looseness=-1

In this work, we investigate the surprisal--duration trade-off by analysing a massively multi-lingual dataset of more than 600 languages \citep{salesky-etal-2020-corpus}. 
We present an experimental framework, controlling for several possible confounds,
and evaluate the surprisal--duration trade-off at the phone level.
We find evidence of a trade-off across
languages: 
languages with more surprising phones compensate by making utterances longer. 
We also confirm mono-lingual trade-offs in 319 languages, out of 600;\footnote{The original number of languages was 635, but after removing those with quality issues, we end up with 600. This process is explained in \cref{sec:data}.}
within these languages, more surprising phones are pronounced with a significantly longer duration.
This is the most representative evidence of the uniform information density hypothesis to date.
Moreover, we did not find evidence of a single language where the opposite effect is in operation (i.e. where more informative phones are shorter).
Given these collective results, we conclude there is strong evidence for a surprisal--duration trade-off both across and within the world's languages.\looseness=-1

\section{Surprisal and Duration}

Cross-linguistic comparisons of information rate go back at least 50 years.
In a study comparing phonemes per second,
\citet{osser1964cross} found no statistical difference between the speech rate of English and Japanese native speakers. 
In a similar study, \citet{denOs1985perception,den1988rhythm} compared Dutch and Italian and found 
no difference in terms of syllables per second, although Italian was found to be somewhat slower in phones per second. 
Such cross-linguistic comparisons, however, are not straightforward, since the range of speech rate can vary widely within a single language, depending on sentence length \citep{fonagy1960speed} and type of speech \citep[e.g. storytelling vs interview;][]{kowal1983use}.
In a meta-analysis of these studies, \citet{roach1998some} concludes that carefully assembled speech databases would be necessary to answer this question.
In this line, \citet{pellegrino2011cross} recently analysed the speech rate of 8 languages using a semantically controlled corpus. They found strong evidence towards non-uniform speech rates across these languages.\looseness=-1

This result is not surprising, however, given that natural languages vary widely in their phonology, morphology, and syntax.
Despite these differences, researchers have hypothesised that there exist compensatory relationships between the complexity of these components \cite{hockett1958course,martinet}.  
For instance, a larger phonemic diversity could be compensated by shorter words \cite{moran2014cross,pimentel-etal-2020-phonotactic} or a larger number of irregular inflected forms could lead to less complex morphological paradigms \citep{cotterell2019complexity}.
Such a compensation can be thus seen as a type of balance, where languages compromise reliability versus 
effort in communication \citep{zipf,martinet1962functional}.
One natural avenue for creating this balance would be a language's information rate. If this were kept roughly constant, the needs of both speakers (who prefer shorter utterances) and listeners (who value easier comprehension) could be accommodated.
Speech rate would then be compensated by information density, resulting in a form of surprisal--duration trade-off.
Indeed, \citet{pellegrino2011cross} and \citet{coupe2019different} present initial evidence of this trade-off across languages.\looseness=-1

Analogously, the UID hypothesis posits that, within a language, users balance the amount of information per linguistic unit with the duration of its utterance.
This hypothesis has been used to explain a range of experimental data in psycholinguistics, including syntactic reduction \cite{levy2007speakers} and contractions, such as \textit{are} vs \textit{'re} \cite{frank2008speaking}.
While this theory is somewhat under-specified with respect to its causal mechanisms,  as we argue in \citet{meister+al.emnlp2021a}, one of its typical interpretations is that users are maximising a communicative channel's capacity \citep{frank2008speaking,piantadosi2011word}.
If we assume this channel's capacity to be constant across languages, we may derive a cross-linguistic version of UID. 
Such a hypothesis would predict, for instance, that speakers of languages with less informative phones will make
them faster.\clara{still don't love that we're citing UID and then using a linear relationship for evidence, which is what we call people out for doing... Can we qualify this a bit more? like saying that this doesnt directly test the implications of the hypothesis?}\tiago{We test exactly the implications here (the trade-off is an implication), right? What we dont test the causes. But we were very clear in the previous sentences that uid is undefined with respect to its cause. We can hedge it more if you want, though... Feel free to give it a shot, I can undo/rewrite if I dislike.}
Under this specific interpretation, our study can be seen as evidence of UID as a cross-linguistic phenomenon.\looseness=-1

\section{Measuring Surprisal} \label{sec:surprisal}

To formalise our approach, we first present a standard measure of information content: \defn{surprisal}. 
In the context of natural language, surprisal \cite{hale2001probabilistic} measures the Shannon information content a linguistic unit conveys in context, which can be measured as its negative log-probability:
\begin{equation}\label{eq:surprisal}
    \ent(S_t = s_t \mid \bS_{<t} = \bs_{<t}) = - \log p(s_t \mid \bs_{<t})
\end{equation}
In this equation, $\bS$ is a sentence-level random variable, with instances $\bs \in \calS^*$, and $t$ indexes a position in the sentence. 
Accordingly, we define $\calS$ as the set of phones in a given phonetic alphabet, and we use $\bs_{<t}$ to indicate the context in which phone $s_t$ appears.\looseness=-1

Unfortunately, this surprisal is not readily available, since we would need access to the true distribution $p(s_t \mid \bs_{<t})$ to compute it.
We will use an approximation $\ptheta(s_t \mid \bs_{<t})$ instead, i.e. a phone-level model with estimated parameters $\btheta$.\looseness=-1

\subsection{Approximating $p(s_t \mid \bs_{<t})$.}
While much of the original psycholinguistic work on surprisal estimated $\ptheta$ using $n$-gram models \citep[\emph{inter alia}]{levy2007speakers,coupe2019different}, recent work has shown that a language model's psychometric
predictive power correlates directly with its quality, measured by its cross-entropy in held-out data \citep{goodkind2018predictive,wilcox2020predictive}.
We will thus make use of LSTMs in this work, since they have been shown to outperform $n$-grams on  phone-level language modelling tasks \citep{pimentel-etal-2020-phonotactic}.
We first encode each phone $s_t$ into a high-dimensional lookup embedding $\be_t \in \R^{d_1}$, where $d_1$ is its embedding size.
We then process these embeddings using an LSTM \citep{hochreiter1997long}, which outputs contextualised hidden state vectors:
\begin{equation}
    \bh_t = \lstm(\bh_{t-1}, \be_{t-1}) \in \R^{d_2}
\end{equation}
where the initial hidden state $\bh_0$
is the zero vector and the initial phone $s_0$ is a start-of-sentence symbol.
The hidden states are then linearly transformed and projected onto $\Delta^{|\calS| + 1}$, the probability simplex, via a softmax to compute the desired distribution:\footnote{The dimension of the probability simplex is $|\calS|+1$ to account for an end-of-sentence symbol.}\looseness=-1%
\begin{equation}
    \ptheta(s_t \mid \bs_{<t}) = \softmax(\bW \bh_t + \bb)
\end{equation}
where $\bW \in \R^{(|\calS|+1) \times d_2}$ and $\bb \in \R^{(|\calS|+1)}$ are learnable parameters.
We optimise the parameters by minimising our model's cross-entropy with a training set, which corresponds to minimising the following objective
\begin{equation}
    \enttheta(S_t \mid \bS_{<t}) = -\sum_{n=1}^{N} \sum_{t=1}^{|\bs^{(n)}|} \log \ptheta(s_t^{(n)} \mid \bs_{<t}^{(n)})
\end{equation}
where we assume $\{\bs^{(n)}\}_{n=1}^{N}$ are sampled from the true distribution $p(\cdot)$.\looseness=-1

To avoid overfitting to this training set, we then estimate the cross-entropy with a validation set  $\{\hat{\bs}^{(m)}\}_{m=1}^{M}$, where we stop training once this validation cross-entropy stops decreasing.
Note that minimising the cross-entropy is equivalent to minimising the Kullback--Leibler divergence between two distributions.
Further, if we have access to a larger number of samples $M$, we assume this cross-entropy estimate will give us a tight approximation to the cross-entropy between $\ptheta$ and the true distribution.
Thus, the lower this cross-entropy, the closer we may assume our model is to the true $p(\cdot)$, and the better we should expect our surprisal estimates to be.\looseness=-1

\paragraph{Hyper-parameter choices.}\clara{shouldn't this go in the experiments and results section?}\tiago{I like it here, since it encapsulates how we get the surprisal. We dont mention it anymore later in the paper.}
We implement our phone-level LSTM language models with two hidden layers, an embedding size of 64 and a hidden size of 128.
We further use a dropout of $0.5$ and a batch size of 64.
We train our phone-level LSTM models using AdamW \citep{loshchilov2018decoupled} with its default hyper-parameters in PyTorch \citep{pytorch}. 
We evaluate our models on a validation set every 100 batches, stopping training when we see no improvement for five consecutive evaluations.
We split each language's data (described in \cref{sec:data}) into train-dev-test sets using an 80-10-10 split, using sentences as our delimiters. We thus do not separate phone data points from the same sentence. 
We use the first two splits to train and validate our models, while the test set is held out and used throughout our analysis.\looseness=-1

\section{Data} \label{sec:data}

\begin{figure}
    \centering
    \includegraphics[width=\columnwidth]{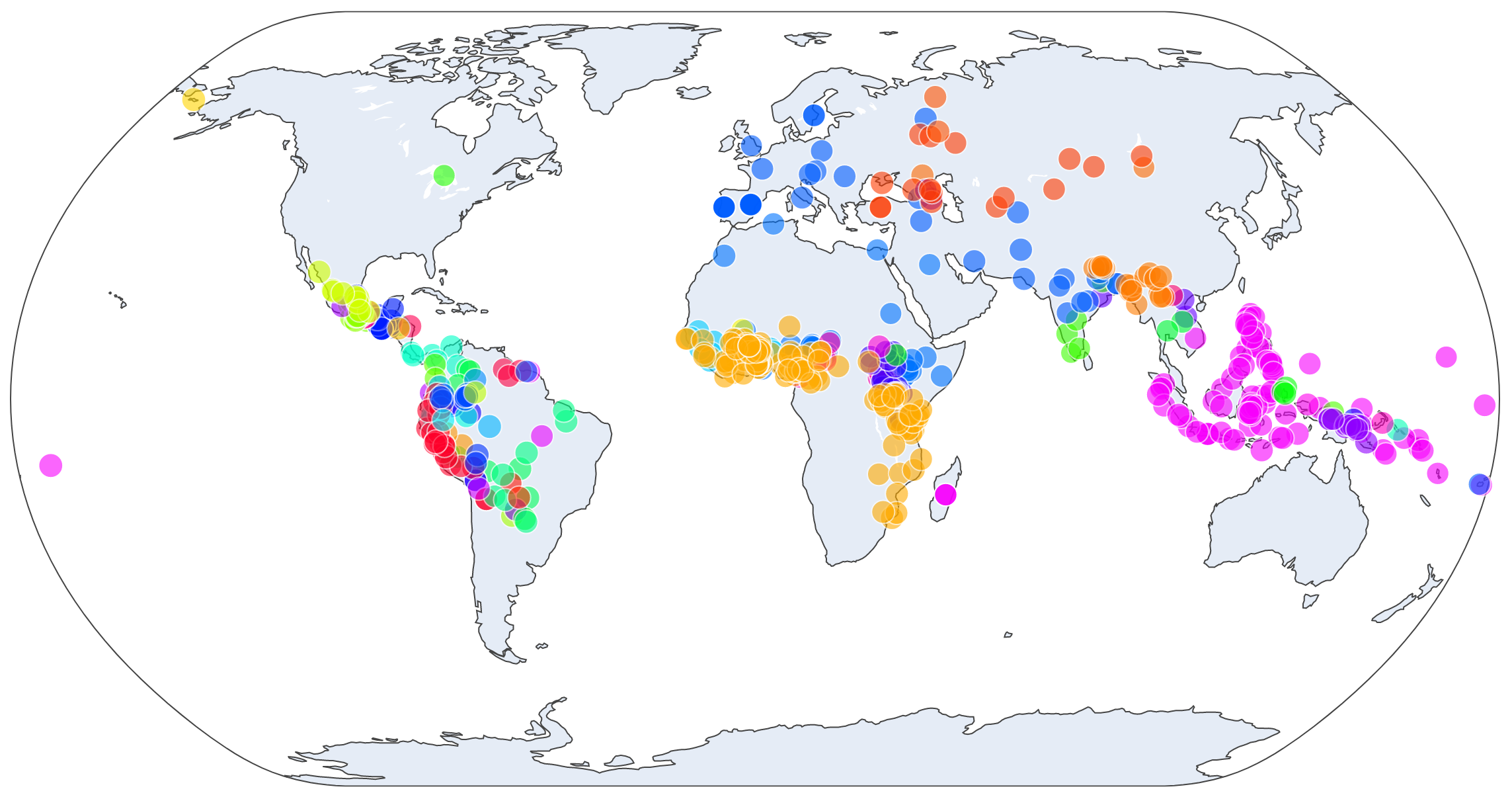}
    \caption{The languages of the VoxClamantis corpus geo-located and coloured by language family.}
    \label{fig:languages_map}
\end{figure}

We use the VoxClamantis dataset for our analysis \citep{salesky-etal-2020-corpus}. 
This dataset is derived from spoken readings of the Bible\footnote{These texts were crawled from \url{bible.is} and utterance-aligned by \citet{black2019wilderness} for the CMU Wilderness dataset.}  and spans more than 600 languages from 70 language families, as shown in \cref{fig:languages_map}.\footnote{A list of all languages can be found in \cref{app:language_list}.}
This dataset offers us a semantically controlled setting for our experiments, as it is composed of translations of a single text, the Bible.

This dataset contains automatically generated phone alignments and derived phonetic measures for all its languages (with both phone duration, and vowels' first and second formant frequencies). 
On average, there are approximately 9,000 utterances (or 20 hours of speech) per language, making it the largest dataset of its kind.
Phone labels were generated using grapheme-to-phoneme (G2P) tools and time aligned using either multilingual acoustic models \cite{wiesner2019asru,povey2011kaldi} or language-specific acoustic models \cite{black2019wilderness,anumanchipalli2011festvox}. 
VoxClamantis offers its phonetic measurements under three G2P models, which trade-off language coverage and quality. 
We will focus on two:\footnote{We set Wikipron \citep{lee2020wikipron} alignments aside because we could not obtain word position information for them.}\looseness=-1
\begin{itemize}%
    \item \defn{Epitran \citep{epitran2018mortensen}.} This is a collection of high quality G2P models based on language-specific rules. Phonetic measurements produced with Epitran are available for a collection of 39 doculects\footnote{The term doculect refers to a dialect as recorded in a specific document, in this case a Bible reading.} from 29 languages (as defined by ISO codes) in 8 language families.
    \item \defn{Unitran \citep{qian2010unitran}.} This is a na\"ive and deterministic G2P model, but its derived measurements are available for all languages in VoxClamantis. While Unitran is particularly error-prone for languages with opaque orthographies \citep{salesky-etal-2020-corpus}, we filter out the languages with lower-quality alignments (as we detail below).
    The original dataset has 690 doculects from 635 languages in 70 language families.
\end{itemize}

In order to study the trade-off hypothesis we require two measurements: phone durations and phone-level surprisals. As mentioned above, phone durations are readily available in VoxClamantis. 
Phone-level surprisals, on the other hand, are not, so we employed phone-level language models in order to estimate them (as detailed in \cref{sec:surprisal}). 
Given both these values, we can now perform our cross-linguistic analysis.
First, though, we will describe some data quality checks.

\paragraph{Filtering Unitran.}
The phone and utterance alignments for the VoxClamantis dataset were automatically generated and may be noisy due to both of these processes. 
The labels from the Unitran G2P also contain inherent noise due to their deterministic nature. 
Accordingly, we filter the data using the mean Mel-Cepstral Distortion (MCD) as an implicit quality measure for the alignments. 
MCD is an edit-distance metric which evaluates the distance between some reference speech and speech synthesised using the alignments \cite{kubicheck1993mcd}.
We use the utterance-level MCD scores from the CMU Wilderness dataset \citep{black2019wilderness}, removing all utterances with an MCD score higher than~7.
This leaves us with 647 doculects from 600 languages in 69 language families.

\section{Design Choices} \label{sec:design_choices}

There are several critical design choices that must be made when performing a cross-linguistic analysis of this nature.
While some may at first seem inconsequential, they can have a large impact on down-stream results. 
Specifically, we assume that there is a surprisal--duration trade-off which is caused by a capacity to process information, which should be roughly constant across human populations.
We must thus control for other potential sources for this trade-off, which we deem to be uninteresting in this work.

\paragraph{Phone-level Analysis.}
While there are good reasons for performing this analysis at the syllable- or word-level, we believe phones are advantageous for out study.
\citet{greenberg1999speaking}, for instance, shows syllables are less prone than phones to be completely deleted in casual speech;  syllables would thus allow more robust estimates of speech duration.
Nonetheless, languages that allow for more complex (and long) syllabic structures will naturally have more valid syllables. 
A larger number of syllables, in turn, will cause each syllable to  be less predictable on  average.\footnote{Probabilities must sum to 1. This finite probability mass means average probability must go down with more classes.}
Therefore, more complex syllables will be both longer and unpredictable. Studying syllables can thus lead to trivial trade-offs which mainly reflect the methodology employed.
A similar argument can be made against word-level analyses.\footnote{Concatenative languages, for instance, would have both longer and less predictable words. Take the German word \emph{Hauptbahnhof} which can be  translated into English as \textit{central train station}. Predicting this single (and long) German word is equivalent to predicting three words in English.\looseness=-1}
Performing this type of analysis at the phone-level should alleviate this effect, making it the more appropriate choice.

\paragraph{Articulatory Costs.}

Whereas the range of effort used to produce individual phones may be smaller than in other linguistic hierarchies, there is still a considerable variation in the cost associated with each phone's articulation.
For instance, \citet{zipfpsycho} argued that a phone's articulatory effort was related to its frequency.
If this is indeed the case, a direct analysis of surprisal--duration pairs that does not control for articulatory effort could also lead to a trivial trade-off: the long and effortful phones will be less frequent and likely to be more unpredictable, having higher surprisals.
To account for each phone's articulatory cost, we use mixed effects models in our analysis, and include phone identity as random intercept effects.

\paragraph{Word-initial and Word-final Lengthening.}

There is ample evidence showing that, across languages, word-initial and word-final segments are lengthened during production \citep{fougeron1997articulatory,white2020initial}.
Another property is that word-initial positions carry more information than word-final ones, which has been well-studied in both psycholinguistics and information-theory.
From a psycholiguistic perspective, it seems word-initial segments are more important for word recognition \citep{bagley1900apperception,fay1977malapropisms,bruner1958note,nooteboom1981lexical}.
Under an information-theoretic analysis, it has been observed that earlier segments in a word are more surprising than later ones \citep{son2003efficient,king2020greater,pimentel-etal-2021-disambiguatory}.
Word-initial segments are both lengthened and more surprising, potentially for unrelated reasons.
An analysis which does not control for such word-positioning is thus doomed to find trivial correlations. 
To account for this word-initial and word-final lengthening, we include three word position fixed effects (initial, middle, or final) in our mixed effects models.

\paragraph{Sentential Context.} 

The amount of context that a model conditions on when estimating probabilities will undoubtedly have an impact on a study of this nature. 
For example, a model that cannot look back beyond the current word, such as the one employed by \citet{coupe2019different}, can by definition only condition on the previous phones in the same word.
Arguably, a cognitively motivated surprisal--duration trade-off should estimate surprisal using a phone's entire sentential context and not %
only  the prior context inside a specific word.
In this work, we make use of LSTMs (as described in \cref{sec:surprisal}), which can model long context dependencies \citep{khandelwal-etal-2018-sharp}.\looseness=-1

\section{Generalised Mixed Effects}

Throughout our experiments, we will use mixed-effects models; we provide a brief introduction here (see \citet{wood2017generalized} for a longer exposition).
Classical linear regressions models can be written as:\looseness=-1
\begin{equation}\label{eq:simple_regression}
    y_i = \bphi^\intercal\,\bx_i + \epsilon_i, \quad \epsilon_i \sim \normal(0, \sigma_{\mathtt{err}}^2)
\end{equation}
where $y_i$ is the target variable, $\bx_i \in \R^{d}$ is the model's input and $\bphi \in \R^{d}$ a learned weight vector.
Further, the error (or unexplained variance) term $\epsilon_i$ is assumed to be normally distributed and independent and identically distributed (i.i.d.) across data instances.
Such an i.i.d. assumption, however, may not hold. 
In our analysis, for instance, multiple phones come from each of our analysed languages; it is thus expected that such co-language phones share dependencies in how their $\epsilon_i$ are sampled.
Mixed-effects models allow us to model such dependencies through the use of random effects.
Formally, for an instance $\bx_{i}$ from a specific language $\lang$, we model:
\begin{equation}\label{eq:mixed_intercept}
    y_{i} = \bphi^\intercal\,\bx_{i} + \omega_{\lang} + \epsilon_i, \quad 
    \begin{array}{c}
         \omega_{\lang} \sim \normal(0, \sigma_{\omega}^2) \\
         \epsilon_i \sim \normal(0, \sigma_{\mathtt{err}}^2)
    \end{array} 
\end{equation}
where $\omega_{\lang}$ is a random effect and $\bphi$ is now termed a fixed effect. 
Here, $\omega_{\lang}$ is an intercept term which is assumed to be shared across all instances of language $\lang$, and $\sigma_{\omega}^2$ is directly learned from the data. 
Similarly, we can add random slope effects:
\begin{equation} \label{eq:mixed_slope}
    y_{i} = \bphi^\intercal\,\bx_{i} + \bbeta_{\lang}^\intercal\,\bx_{i} + \omega_{\lang} + \epsilon_i,\, 
    \begin{array}{c}
         \bbeta_{\lang} \sim \normal(0, \Sigma_{\bbeta}) \\
         \omega_{\lang} \sim \normal(0, \sigma_{\omega}^2) \\
         \epsilon_i \sim \normal(0, \sigma_{\mathtt{err}}^2)
    \end{array} 
\end{equation}
where each $\bbeta_{\lang}\in\R^d$ is a language-specific random slope and $\Sigma_{\bbeta}$ is a (learned) covariance matrix.
Furthermore, our assumption that error terms are normally distributed may not hold in this setting. 
Phone durations, for instance, cannot be negative and are positively skewed, making a log-linear model more appropriate:
\begin{equation} \label{eq:mixed_generalized}
    \log(y_{i}) = \bphi^\intercal\,\bx_{i} + \bbeta_{\lang}^\intercal\,\bx_{i} + \omega_{\lang} + \epsilon_i
\end{equation}
where $\bbeta_{\lang}$, $\omega_{\lang}$, and $\epsilon_{i}$ are still distributed as in \cref{eq:mixed_slope}.
This is similar to modelling the original $\epsilon_{i}$ terms as coming from a log-normal distribution.
We note though, that under this model our effects become multiplicative (as opposed to additive): an increase of $\delta$ unit in the right side will make the value of $y_{i}$ be multiplied by $e^{\delta}$. 
We will use \texttt{lme4}'s \citep{lme4} notation to represent these models. 
Under this notation, a parenthesis represents a random effect and parameters are left out. 
We thus re-write \cref{eq:mixed_generalized} as:
\begin{equation}
    \log(y) = 1 + \bx + (1 + \bx \mid \mathtt{language})
\end{equation}

\section{Experiments and Results\footnote{Our code is available at \url{https://github.com/rycolab/surprisal-duration-tradeoff}.}}

In this section, we will first analyse the surprisal--duration trade-off in individual languages.
We will then perform an analysis with our full data, studying the trade-off both within and across languages with a single model. 
Finally, in our last experiment we will average phone information per language to analyse a purely cross-linguistic trade-off.

\subsection{Individual Language Analyses}

We first analyse languages individually, verifying if more surprising phones have on average a longer duration.
With this in mind, we estimate a generalised mixed effects model for each language. 
We control for each phone's articulatory costs by adding phone identity as a random effect. 
Additionally, we include fixed effects to control for word position effects, adding separate intercepts for word-initial and word-final positions. 
Finally, we consider a fixed effect relating surprisal and word positions. 
At word-initial positions, for instance, the connection between surprisal and duration could potentially be stronger or weaker.\footnote{We analyse the impact of both these effects, phone identity and word position, in \cref{sec:word_positions}.}
This leaves us with the following relationship:
\begin{align} \label{eq:mono-lingual-model}
\small
    &\log(\mathtt{duration}) = 
    1 + \mathtt{surprisal} + \mathtt{position}  \nonumber\\
    &\,\,\, +\mathtt{surprisal} \cdot \mathtt{position} + 
    (1 \mid \mathtt{phone}) 
\end{align}
In this parametrisation, a trade-off between surprisal and duration will emerge as a positive and significant \texttt{surprisal} slope.
Analogously, an inverse trade-off will emerge as a negative and significant slope, since we use two-tailed statistical tests.\footnote{Statistical significance was assessed under a confidence level of $\alpha < 0.01$ and we used \citet{benjamini1995controlling} corrections for multiple tests whenever necessary.}
Out of the 39 doculects in Epitran, 30 present statistically significant positive slopes ($\sfrac{23}{29}$ languages, and $\sfrac{8}{8}$ families; meaning at least one language showed a significant effect per family). 
On Unitran (which we recall is a noisier dataset), $\sfrac{326}{647}$ doculects presented significantly positive slopes ($\sfrac{319}{600}$ languages, and $\sfrac{53}{69}$ families). 
Additionally, we find no language in either dataset with significantly negative slopes: we either find evidence for the trade-off or we have no association whatsoever.\looseness=-1

\begin{figure}
    \centering
    \includegraphics[width=\columnwidth]{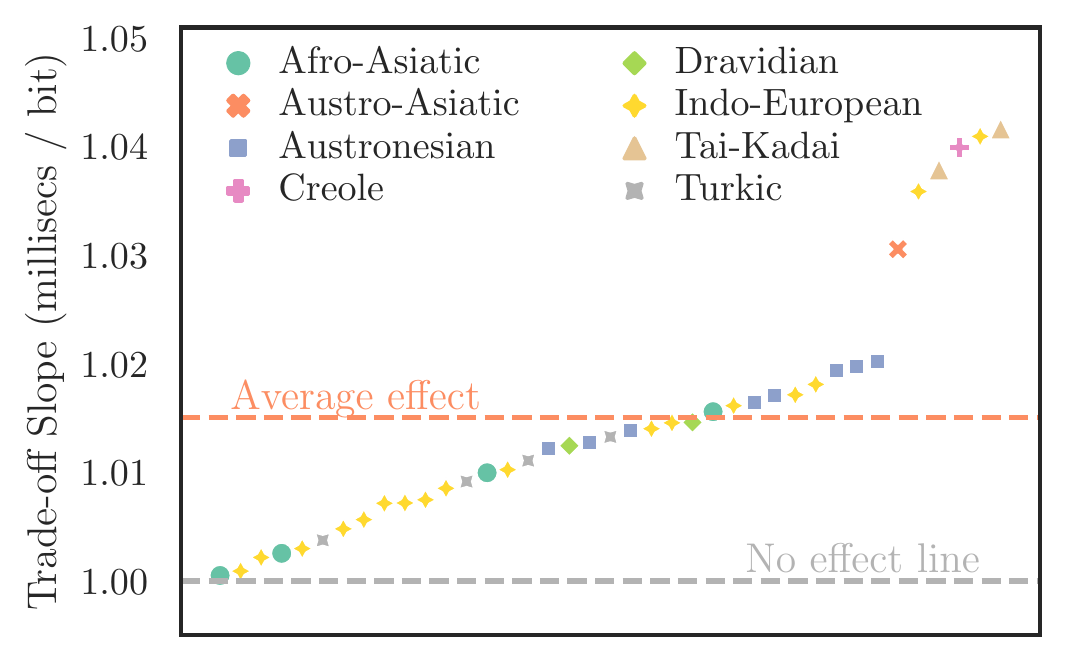}
    \caption{Language-specific trade-off slopes in Epitran from the mixed effects model in \cref{eq:mono-lingual-model}. The $y$-axis represents a multiplicative effect, duration is multiplied by $y$ per extra bit of phone information.}
    \label{fig:monoling_effects}
\end{figure}

The trade-off strength, as measured by the surprisal--duration slopes, can be seen in \cref{fig:monoling_effect_full} (on first page) and \cref{fig:monoling_effects}.
As noted above, by predicting a linear change in logarithmic scale, our effects become multiplicative instead of additive.
The average multiplicative slope we get across all the analysed languages in both datasets is roughly $1.02$, meaning that 
each added bit of information multiplies duration by 1.02.
We believe this should serve as strong support for our hypothesis of a trade-off within languages.
Moreover, to the best of our knowledge, this is the most representative study of the UID hypothesis to date, as measured by the number and typological diversity of analysed languages.\looseness=-1

\subsection{Aggregated Cross-linguistic Analysis} \label{sec:cross_ling_analysis}

\begin{figure}
    \centering
    \includegraphics[width=\columnwidth]{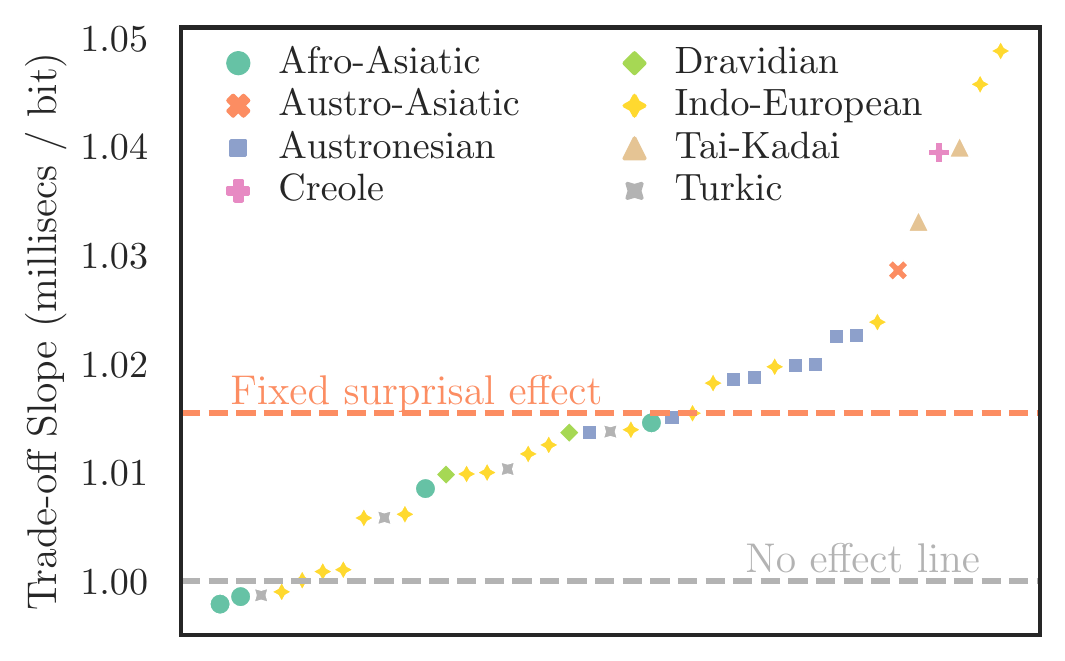}
    \caption{Language-specific trade-off slopes in Epitran from the mixed effects model in \cref{eq:cross-ling-model}. The $y$-axis represents a multiplicative effect, duration is multiplied by $y$ per extra bit of phone information.}
    \label{fig:crossling_effects}
\end{figure}

Following the previous study, we now run a cross-linguistic analysis by aggregating all the languages within a single model. We add the same controls as before, but further nest the phone random effects per language (meaning we create one random effect per phone--language pair). We also include random language-specific intercepts and slopes. 
Formally,
\begin{align} \label{eq:cross-ling-model}
\small
    &\log(\mathtt{duration}) = 
    1 + \mathtt{surprisal} + \mathtt{position} \nonumber \\
    &\quad +\mathtt{surprisal} \cdot \mathtt{position} \nonumber \\
    &\quad +(1 + \mathtt{surprisal} + \mathtt{surprisal} \cdot \mathtt{position} \nonumber \\
    &\qquad\qquad\qquad \mid \mathtt{language}) \nonumber \\
    &\quad +(1 \mid \mathtt{language:phone}) 
\end{align}
After estimating this generalised mixed effects model, we find statistically significant cross-linguistic trade-off effects in both datasets. The multiplicative slope is roughly $1.02$ in both datasets, again meaning each extra bit of information multiplies the duration by this value ($\phi=1.023$ in Unitran and $\phi=1.015$ in Epitran).\footnote{For the Epitran data, we performed this analysis while also adding language family effects and found similar results. However, we could not repeat this experiment for Unitran as the model was too memory intensive.\looseness=-1}
We further analyse the per-language trade-off slopes, which can be seen in \cref{fig:crossling_effects}. 
These language-specific slopes are calculated by summing the fixed effect of the surprisal term with its random effects per language.
We see a similar trend in this figure as in \cref{fig:monoling_effects}, with most of the analysed doculects having a positive surprisal--duration trade-off.

\subsection{Cross-linguistic Trade-offs} \label{sec:cross_ling_tradeoff}

Our previous experiment in \cref{sec:cross_ling_analysis} makes use of language-specific random effects.
These effects allow the model to potentially represent \emph{within}-language trade-off effects, while correcting for cross-linguistic differences by using the model parameters.
It therefore cannot serve as confirmation of a trade-off across the world's languages by itself, only as additional evidence for it.
In this section, we do not use language-specific random effects; instead, we average surprisal within a language for each phone--position tuple.
We then train the following mixed effects model:
\begin{align} \label{eq:cross-ling-trade-off}
\small
    &\mathtt{duration} = 
    1 +  \mathtt{surprisal} + \mathtt{position} \nonumber \\
    &\, +\mathtt{surprisal} \cdot \mathtt{position}  + (1 \mid \mathtt{phone}) 
\end{align}
This equation is identical to the one in \cref{eq:mono-lingual-model}, but now we model the language--phone--position tuples, instead of a language's individual phones.\footnote{We note that phone labels may not always align exactly across languages here, due to possible differences between VoxClamantis' G2P label sets. This may introduce noise into this analysis. It is reassuring, though, that the previous analyses with phones as language-specific effects lead to similar conclusions.\looseness=-1}
Additionally, since we are aggregating results per tuple for this analysis, the central limit theorem tells us our model's residuals should be roughly Gaussian.
We thus use linear mixed effects models, instead of the generalised log-linear ones. 
By analysing this model we find a significantly positive surprisal--duration additive slope, of $\phi=1.5$ milliseconds per bit ($\phi=1.54$ in Unitran and $\phi=1.52$ in Epitran).
This confirms the expected cross-linguistic trade-off: languages with more surprising phones really have longer durations, even after controlling for word positions and phone-specific articulatory costs.\looseness=-1

\section{Discussion}

The pressure towards a specific information rate (potentially set at a specific cognitive channel capacity) has been posited as an invariant across languages. 
Directly testing such a claim is perhaps impossible, as data alone cannot prove its universality.
Moreover, providing meaningful evidence towards this phenomenon
requires a careful and comprehensive cross-linguistic analysis, which we attempt to perform in this work. 
In comparison to similar studies, such as those by \citet{pellegrino2011cross} and \citet{coupe2019different}, we employ more sophisticated techniques to measure a linguistic unit's (in our case, a phone's) information content.  
Moreover, we also employ more rigorous strategies for analysing the surprisal--duration relationship, controlling for several potential confounds. By introducing these improvements, we attain a more detailed understanding of the role of information in language production, both across and within languages.\looseness=-1

Experimentally, we find that, after controlling for other artefacts, the information conveyed by a phone in context has a modest but significant relationship with phone duration.
We see that this relationship is consistently positive across a number of investigated settings, despite being small in magnitude, meaning that more informative phones are on average longer.
Additionally, using two-tailed tests at $\alpha<0.01$  throughout our experiments, we find no language with a significant negative relationship between phone surprisal and duration.

\paragraph{Limitations and Future Work.}
In this work, we implemented a careful evaluation protocol to study the relationship between a phone's surprisal and duration in a representative set of languages.
To perform our study in such a large number of languages, however, we rely on the automatically aligned phone measurements from VoxClamantis, which contain noise from various sources.
Future work could investigate if biases in the dataset generation protocol could impact our results.
Further, VoxClamantis data is derived from readings of the Bible. 
Future studies could extend our analysis to other settings, such as conversational data.

\section{Conclusion}

In this work, we have provided the widest cross-linguistic investigation of phone surprisal and duration to date, covering 600 languages from over 60 language families spread across the globe. 
We confirm a surprisal--duration trade-off both across these analysed languages and within a subset of 319 of them, covering 53 language families.
While there exist arguments against some of our design choices, 
our overarching conclusion is remarkably consistent across our analyses: 
the presence of a surprisal--duration trade-off is significant in language production.
In other words, both across and within languages, phones carrying more information are longer, while phones carrying less information are produced faster.\looseness=-1

\bibliography{main}
\bibliographystyle{acl_natbib}

\clearpage
\appendix

\section{Confound analysis} \label{sec:word_positions}

In this section, we analyse the word position and articulatory cost confounds mentioned in \cref{sec:design_choices}, as they could have an impact on a surprisal--duration trade-off analysis.
We first investigate the parameters from our mixed effects models containing word positioning effects.
The word-initial and word-final intercepts are significantly positive in all 39 languages of our mono-lingual Epitran analysis (represented by \cref{eq:mono-lingual-model}) and in both cross-linguistic experiments (\cref{eq:cross-ling-model,eq:cross-ling-trade-off}).
The intercepts for word-initial positions average at $67$ milliseconds, while the word-final ones average at $32$, providing new evidence for this word boundary lengthening effect.
Since word position is correlated with surprisal, this boundary lengthening phenomenon could pose as a source of bias in our results, had we not controlled for it.

We now explore the potential bias introduced by phone-specific articulatory costs. 
As mentioned in \cref{sec:design_choices}, languages with larger phonetic inventory sizes may be more inclined to use marked phones, which have longer duration.
While this correlation between inventory size and unit cost would be particularly problematic for larger linguistic units (e.g. syllables) it can also affect our phone-level analysis.
In fact, we take the Spearman correlation between a language's inventory size (in number of unique phones) and its average phone duration, finding a positive correlation of $\rho=.28$. The average surprisal--duration Spearman correlation across languages is $\rho=0.45$.
As inventory size and surprisal are strongly correlated across languages, we find that pure inventory effects may be driving a large part of the analysed correlation.

To analyse how strongly both confounds would reflect in the main effect if left unaccounted for, we rerun our previous analyses, but without effects for either \texttt{position}, \texttt{phone}, or both. We do so for Epitran only.
The resulting estimated trade-off effects  are given in \cref{tab:confounds}.
We indeed see that these confounds are typically absorbed by the fixed surprisal effect in all three settings.
Notably, without confound control we would find supposedly significant results in all analysed languages, and a 10 times stronger cross-linguistic effect, all of which are in fact spurious.\looseness=-1

\newpage

\begin{table}[H]
    \centering
\resizebox{\columnwidth}{!}{%
    \begin{tabular}{cccccrr}
        \toprule 
        && \multicolumn{4}{c}{Trade-off Slope $\phi$} \\ 
        \cmidrule(lr){3-6}
        \multicolumn{2}{c}{Controls} & \multicolumn{2}{c}{Mono-lingual} & \multicolumn{2}{c}{Cross-linguistic} \\
        \cmidrule(lr){1-2} \cmidrule(lr){3-4} \cmidrule(lr){5-6}
        Phone & Position & \cref{eq:mono-lingual-model} & \# Sign & \cref{eq:cross-ling-model} & \cref{eq:cross-ling-trade-off} \\
        \midrule
        \cmark & \cmark & 1.02 & 30 & 1.02$^\ddagger$ & 1.52$^\ddagger$ \\
        \cmark & \xmark & 1.02 & 37 & 1.03$^\ddagger$ & 0.93$^\ddagger$ \\
        \xmark & \cmark & 1.03 & 33 & 1.02$^\ddagger$ & 15.75 \\
        \xmark & \xmark & 1.04 & 39 & 1.04$^\ddagger$ & 15.73$^\dagger$ \\
        \bottomrule
    \end{tabular}
}
    \caption{Comparison of trade-off (in milliseconds per bit) found when not conditioning on potential confounds. \# Sign represents the number of significant languages ($\alpha<0.01$) in a mono-lingual analysis.}
    \label{tab:confounds}
\end{table}

\section{Additive Effects} \label{app:additive_effects}

\begin{figure}[H]
    \centering
    \includegraphics[width=\columnwidth]{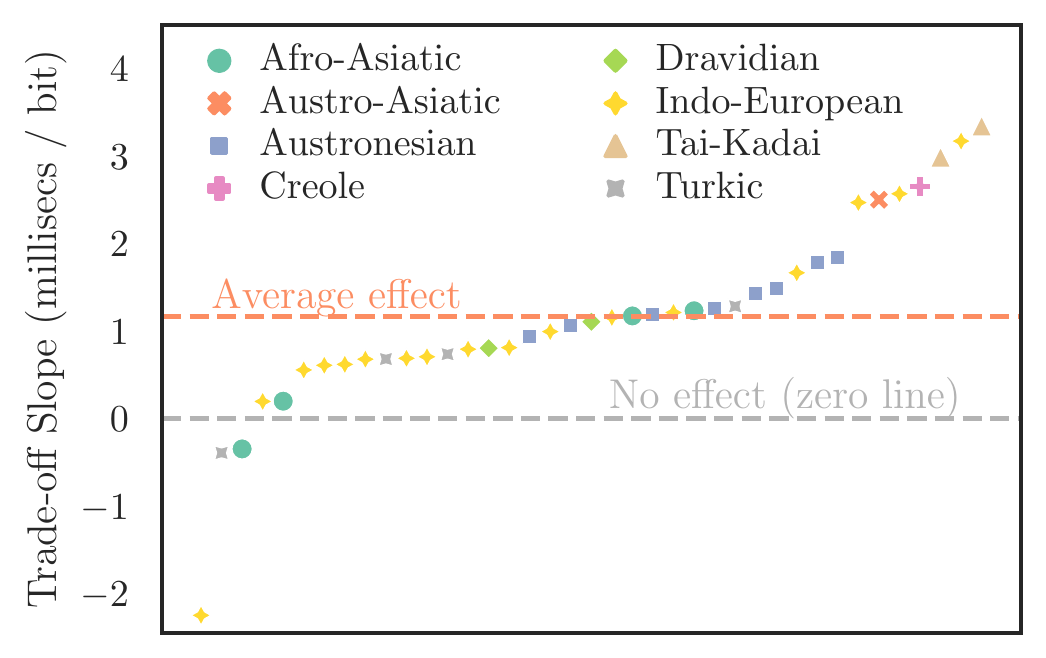}
    \caption{Additive slope of the model in \cref{eq:mono-lingual-model}.}
\end{figure}

\begin{figure}[H]
    \centering
    \includegraphics[width=\columnwidth]{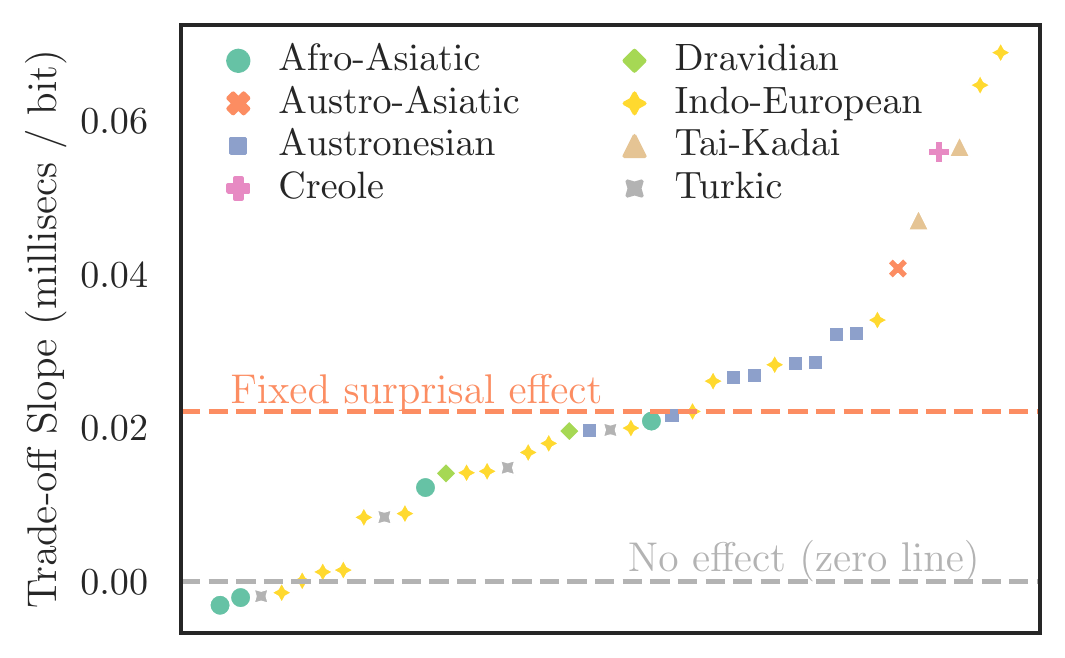}
    \caption{Additive slope of the model in \cref{eq:cross-ling-model}.}
\end{figure}

\hfill
\clearpage

\pagebreak
\onecolumn

\section{Languages}
\label{app:language_list}

The languages used in our analyses are listed below, grouped by language family, along with their three character ISO 639-3 code, and the grapheme-to-phoneme schemes for which phone alignments are available for that language in the VoxClamantis dataset -- Unitran: \textbf{U}, Epitran: \textbf{E} \cite{salesky-etal-2020-corpus}. 
ISO codes for which there are multiple languages listed may represent dialects or other sub-language variations and/or multiple available Bible versions for which data is available.\\

\centerline{\includegraphics[width=0.92\textwidth]{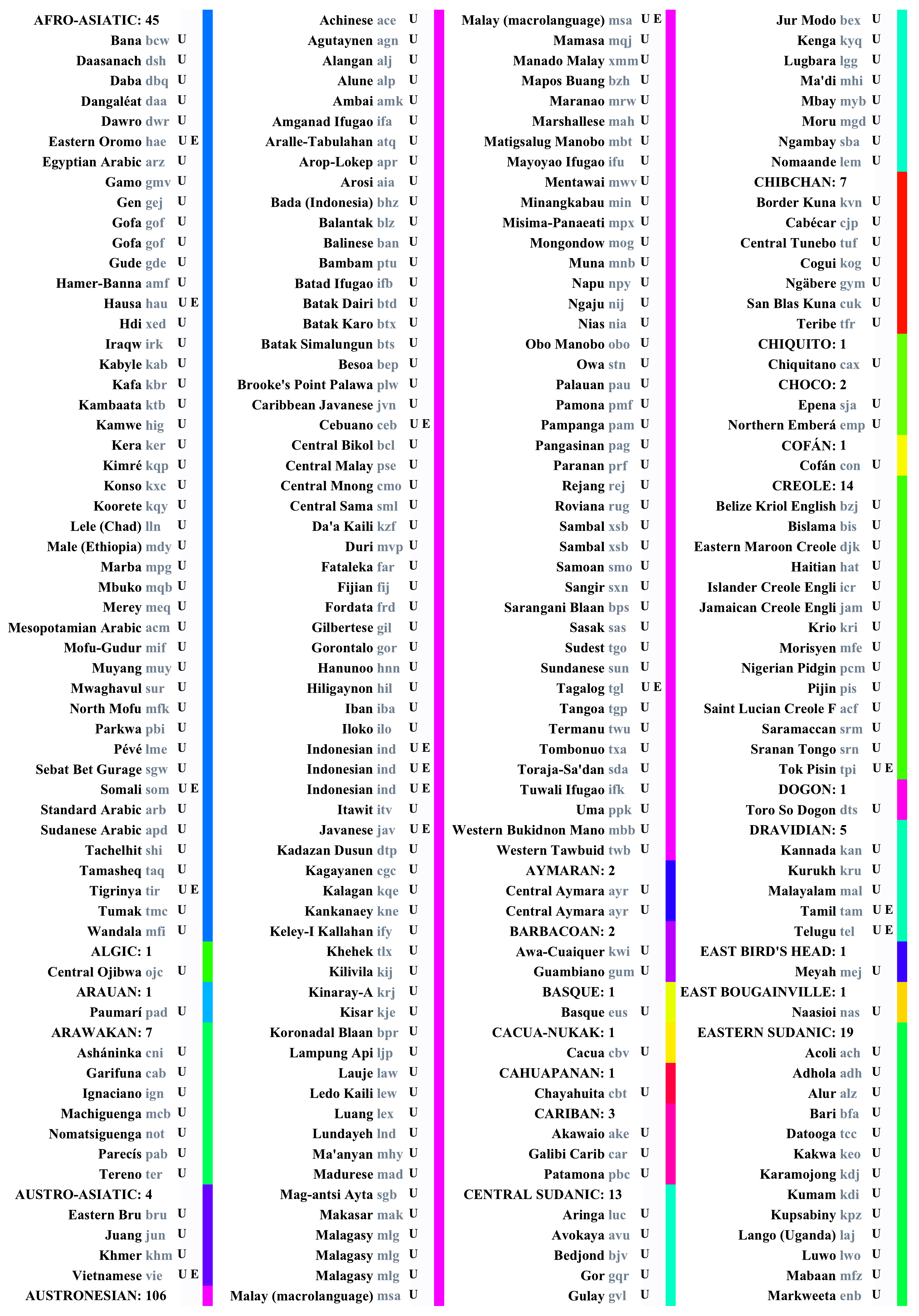}}

\centerline{\includegraphics[width=\textwidth]{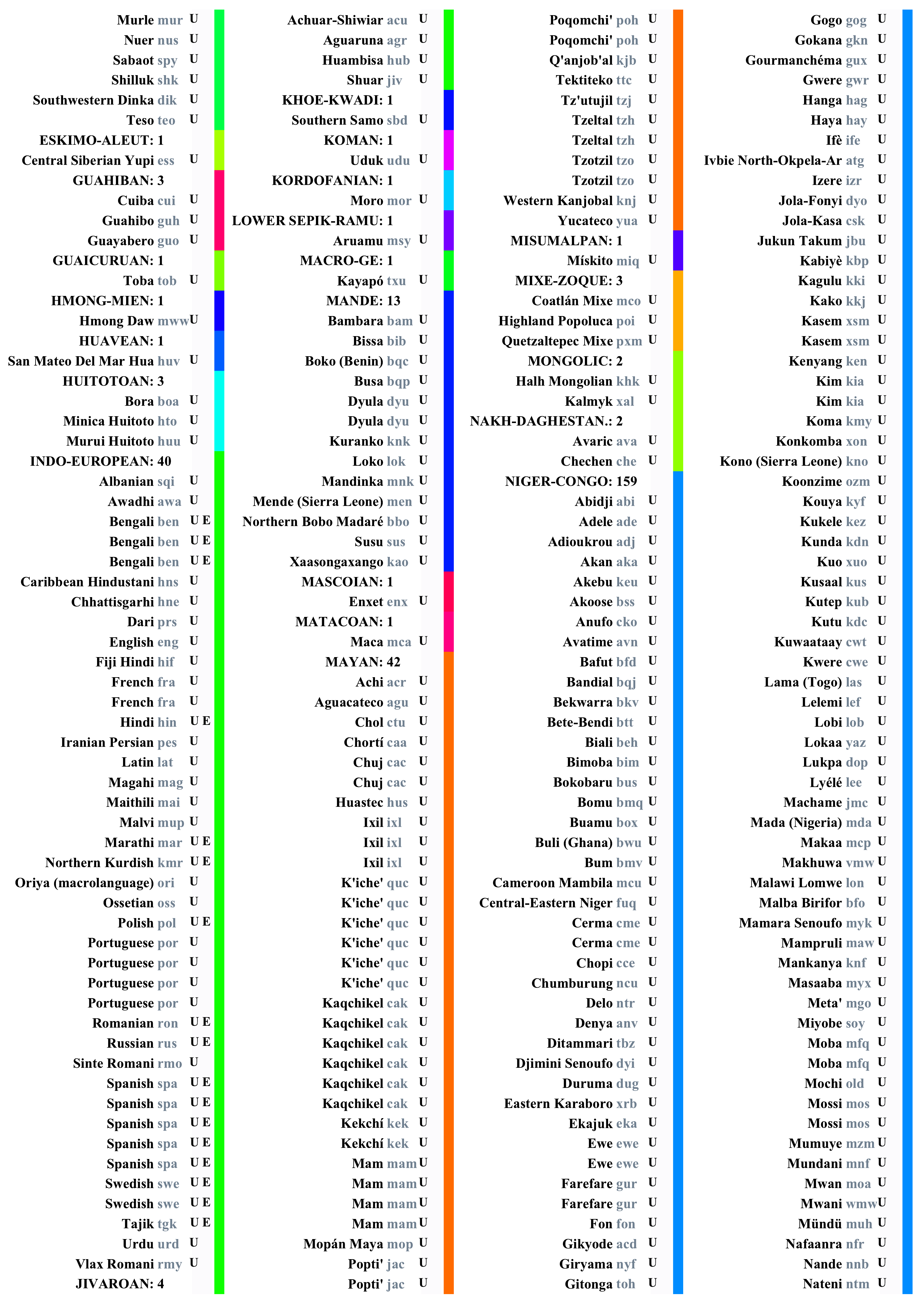}}

\centerline{\includegraphics[width=\textwidth]{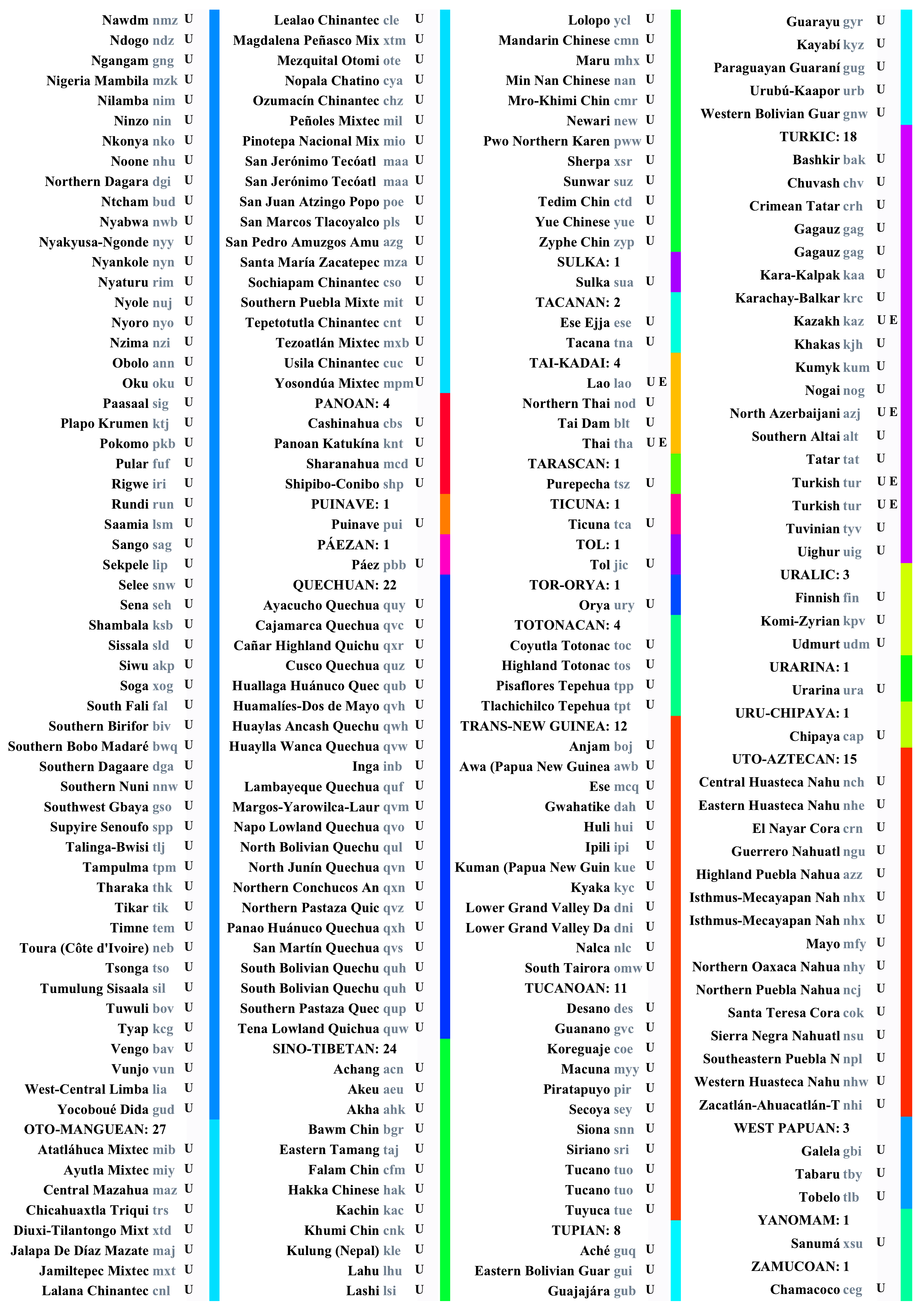}}

\end{document}